\documentclass[conference]{IEEEtran}
\IEEEoverridecommandlockouts
\usepackage{cite}
\usepackage{amsmath,amssymb,amsfonts}
\usepackage{algorithmic}
\usepackage{graphicx}
\usepackage{textcomp}
\usepackage{xcolor}
\def\BibTeX{{\rm B\kern-.05em{\sc i\kern-.025em b}\kern-.08em
    T\kern-.1667em\lower.7ex\hbox{E}\kern-.125emX}}
\begin{document}

\title{Bengali Handwritten Grapheme Classification: Deep Learning Approach

{\footnotesize \textsuperscript{*}Note: pre-print}
}

\author{\IEEEauthorblockN{1\textsuperscript{st} Tarun Roy}
\IEEEauthorblockA{\textit{Dept. of CS} \\
\textit{University of Iowa}\\
Iowa City, USA \\
tarunkanti-roy@uiowa.edu}
\and
\IEEEauthorblockN{2\textsuperscript{nd} Hasib Hasan}
\IEEEauthorblockA{\textit{Dept. of CS} \\
\textit{University of Iowa}\\
Iowa City, USA \\
hasibul-hasan@uiowa.edu}
\and
\IEEEauthorblockN{3\textsuperscript{rd} Kowsar Hossain}
\IEEEauthorblockA{\textit{Dept. of CS} \\
\textit{University of Iowa}\\
Iowa City, USA \\
kowsar-hossain@uiowa.edu}
\and
\IEEEauthorblockN{4\textsuperscript{th} Masuma Akter Rumi}
\IEEEauthorblockA{\textit{Dept. of CS} \\
\textit{University of Iowa}\\
Iowa City, USA \\
masuma-akter@uiowa.edu}
}

\maketitle

\begin{abstract}
Despite being one of the most spoken languages in the world ($6^{th}$ based on population), research regarding Bengali handwritten grapheme (smallest functional unit of a writing system) classification has not been explored widely compared to other prominent languages. Moreover, the large number of combinations of graphemes in the Bengali language makes this classification task very challenging. With an effort to contribute to this research problem, we participate in a Kaggle competition \cite{kaggle_link} where the challenge is to separately classify three constituent elements of a Bengali grapheme in the image: grapheme root, vowel diacritics, and consonant diacritics. We explore the performances of some existing neural network models such as Multi-Layer Perceptron (MLP) and state of the art ResNet50. To further improve the performance we propose our own convolution neural network (CNN) model for Bengali grapheme classification with validation root accuracy 95.32\%, vowel accuracy 98.61\%, and consonant accuracy 98.76\%. We also explore Region Proposal Network (RPN) using VGGNet with a limited setting that can be a potential future direction to improve the performance.
\end{abstract}


\section{Introduction}
Automatic handwritten character recognition (HCR), optical character recognition (OCR) has a lot of commercial and academic interests. However, for these applications Bengali language is much more challenging the English. However, the challenging part is that the Bengali alphabet comprises 50 letters (11 vowels and 39 consonants); there are also 18 diacritics which results in \textasciitilde 13,000 different grapheme variations (compared to English's 250 graphemic units). This large number of combinations of graphemes increases the challenges exceedingly. Moreover, many similar shaped characters exist in the Bengali language. In some cases, the only difference between two similar characters is a single dot or a mark (see figure 1). Therefore, it is difficult to achieve a better performance recognizing Bengali handwritten characters with simple deep learning techniques. This task needs improved techniques and careful consideration of existing challenges. To mitigate this challenge, Bangladesh based non-profit organization $Bengali.AI$ \cite{bengaliAI_link} has built and released crowd-sourced,  metadata-rich datasets \cite{alam2021large} and open sourced them using Kaggle competition \cite{kaggle_link}.

The capability of deep learning techniques to recognize objects from image data has broadened the applicability of techniques to recognize handwritten characters. Recently, deep learning techniques have been used to recognize handwritten characters\cite{kim2014}. However, depending on the language, the challenges and complexity of handwritten character recognition have significant differences. 
     \begin{figure}
     \centering
     \includegraphics[width=0.4\textwidth]{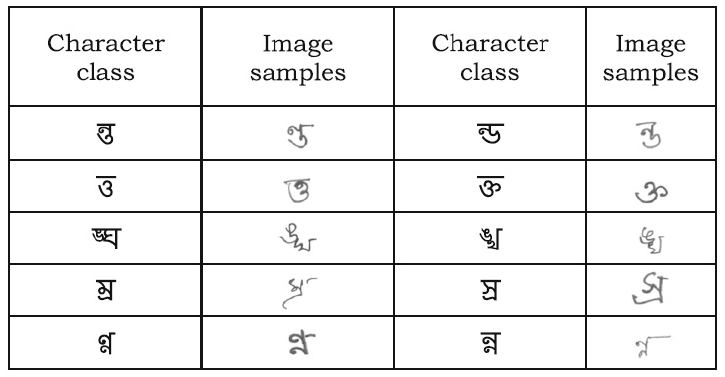}
     \caption{Random samples from a few closely resembling classes \cite{Saikat2007} }
     \label{fig:5}
     \end{figure}

The deep convolutional neural networks (DCNN) have played an important role in the discussion on deep learning recently. The concept of Max Pooling, the introduction of Dropout, etc. led to widespread applications of DCNNs and deep neural networks (DNNs) in general. They have been used for character recognition \cite{journals/corr/abs-1003-0358}\cite{CiresanS13}, object detection, medical imaging among other pattern recognition problems \cite{Hav} etc. DNNs have been proved to be very successful on datasets like MNIST, TIMIT, ImageNet, CIFAR and SVHN, and so on. 

In this project, first we explore the performance of a simple Multi-Layer Perceptron (MLP) model. No wonder the performance is not that good because of the simplicity of the model comparing to the enormous complex dataset. We also explore the performance of ResNet50 on the dataset. Though ResNet50 performs moderately well on our dataset, it crosses the time limit for the training that is set by Kaggle. Therefore, we propose our own model which can train the dataset within the time limit and also acquire good accuracy (more on section 4). Finally, we also try Regional Proposal Network (RPN) method and observe the performance in a limited setting. 

\section{Related Work}
To facilitate our discussion and motivate our work, we give a description of related works in this section. There are significant amount of works on handwritten Bengali character recognition reported through the last decade (\cite{uma_2004}, \cite{subhadip_2012}, \cite{basu_2009}). However, there are limited works on Bengali compound character recognition and grapheme recognition. The accuracy of the system is also a big challenge. 

\cite{pal_2008} explored Bangla handwritten compound character recognition using a modified quadratic discriminant function (MQDF) based on directional information obtained from the arctangent of the gradient. Using a 5-fold cross-validation technique they obtained 85.90\% accuracy from a dataset of Bengali compound characters containing 20,543 samples. In another
work \cite{das_2009}, Quad tree-based features were used for recognition of 55 frequently occurred compound characters in the Bengali language. To do this they used Multi-Layer Perceptron (MLP) classifier. The work presented in \cite{subhadip_2012} involves the design of a MLP based classifier for recognition of handwritten Bengali alphabet using a 76 element feature set. The feature set includes 24 shadow features, 16 centroid features and 36 longest-run features. They observed the accuracy of the system is 86.46\% and 75.05\% on the samples of the training and the test sets respectively. 

Very recent work of \cite{rumman_2019}, explored convolutional neural network (CNN) based Bengali handwritten character recognition system, where the recognition accuracy is 85.36\% on their own dataset for Bengali character recognition. Another recent work \cite{noor_2018} explored Handwritten Bengali Numeral Recognition Using Ensembling of CNN with good accuracy. Good results recognizing Bengali characters and numerals have indicated that CNN can be a good choice to Bengali grapheme classification. However, to the best of our knowledge, there are no significant works that have explored grapheme classification using CNN with high accuracy.

    \begin{figure}
    \centering
    \includegraphics[width=0.4\textwidth]{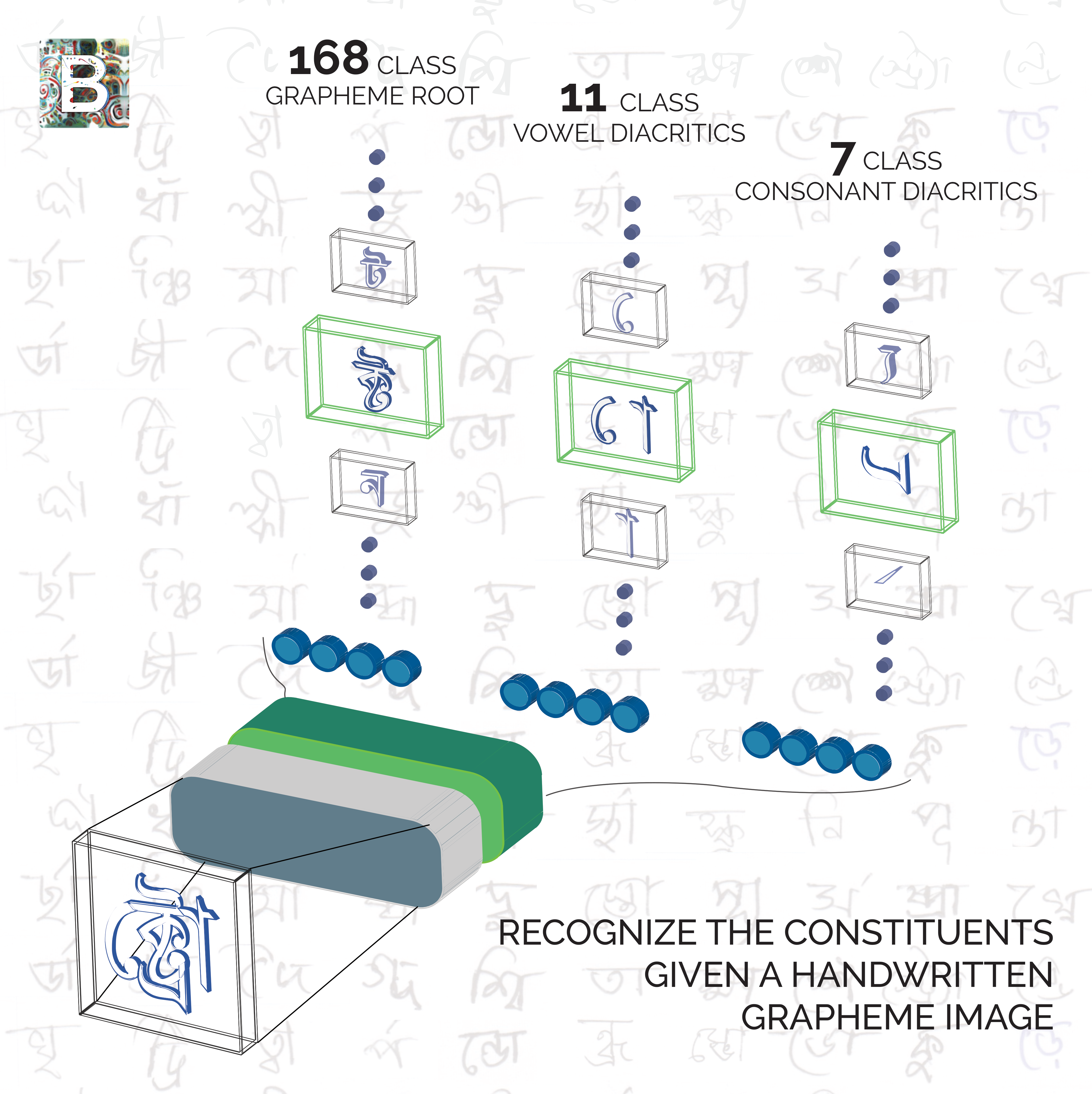}
    \caption{Class map of a grapheme image}
    \label{fig:overview}
    \end{figure}

    \begin{figure}
    \centering
    \includegraphics[width=0.4\textwidth]{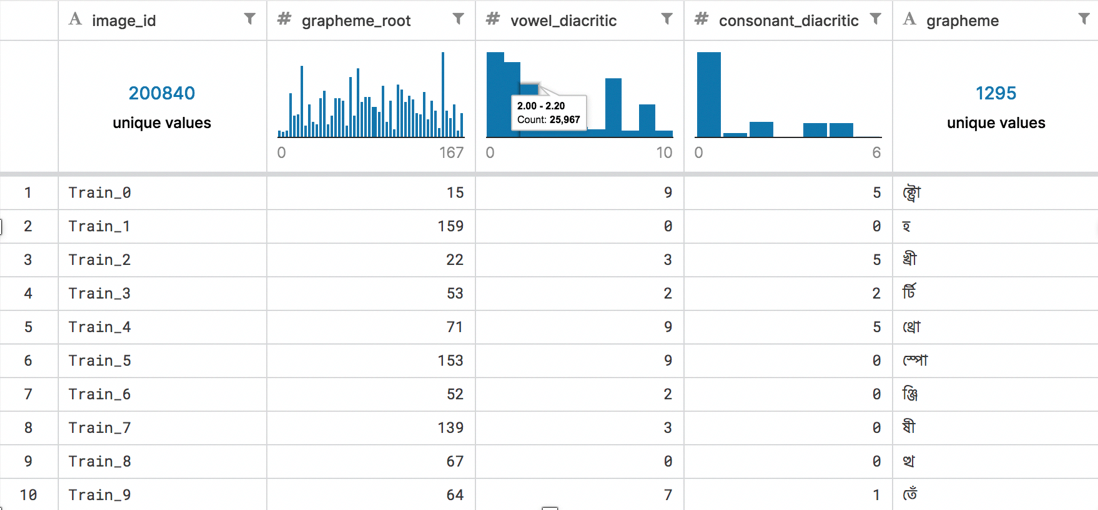}
    \caption{Dataset details}
    \label{fig:dataset}
    \end{figure}
    
\begin{figure*}
\vskip 0.2in
\begin{center}
\centerline {
\begin{tabular}{ccc}
  \includegraphics[width=0.3\textwidth]{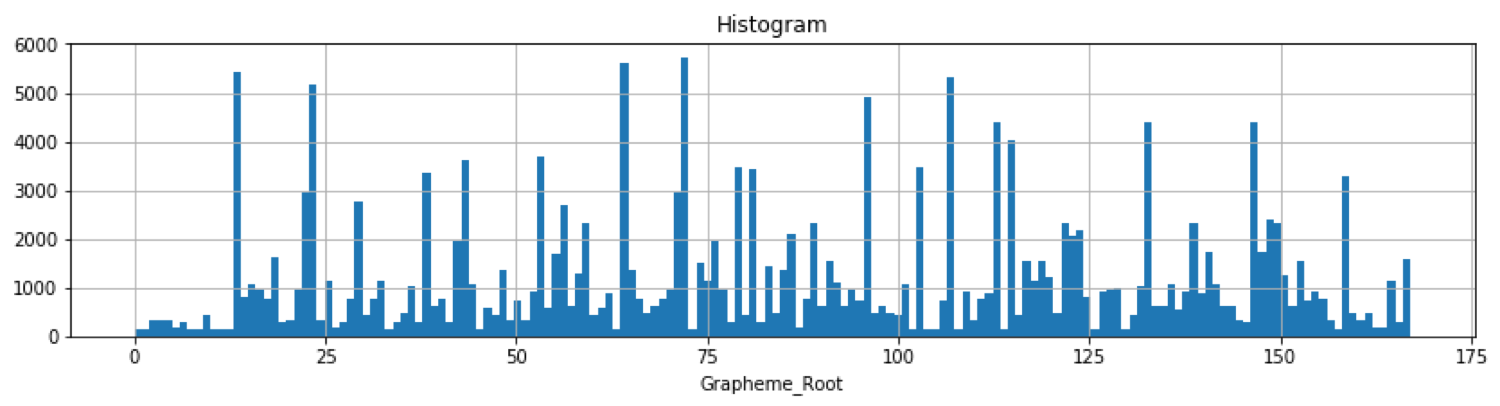} &   \includegraphics[width=0.3\textwidth]{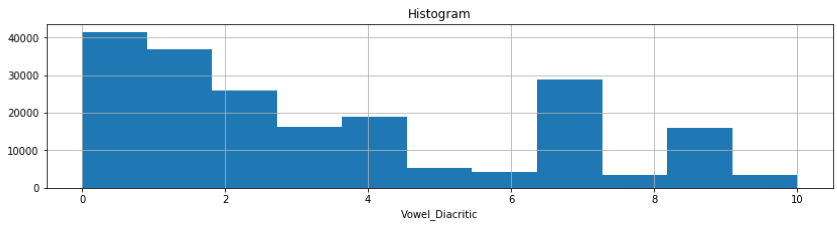} & \includegraphics[width=0.3\textwidth]{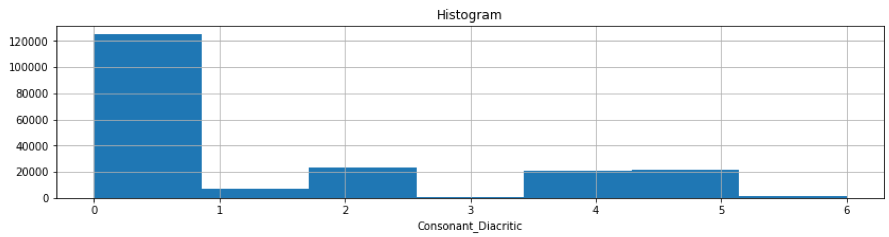}\\
(a) Grapheme roots & (b) Vowel diacritics \& (c) Consonant diacritics\\[6pt]
\end{tabular}
}
\caption{Frequency of different grapheme roots and diacritics}
\label{fig:histogram}
\end{center}
\vskip -0.2in
\end{figure*}

\section{Dataset}
We use Kaggle dataset \cite{kaggle_dataset} for this work which consists of train data, test data, and a class map file that shows the true labels for each image in the train dataset. Train data includes total $200,840$ grapheme images of size $137x236$, partitioned into $4$ separate parquet files. Figure \ref{fig:dataset} shows an overview of the train data. The test data also consists of $4$ separate parquet files that contain about the same amount of image data as the train data. However, test data is not disclosed publicly. For each grapheme in the train dataset, there are three different true labels for root, vowel, and consonant components. The class-map assigns these true labels to every image in the training dataset. Figure \ref{fig:overview} shows the sample class-map for an input grapheme image. In order to get an overview of the dataset, we look into the histogram of the train images from which we observe that the dataset is not properly distributed as shown in Figure \ref{fig:histogram}.
We also notice that some of the combinations of grapheme components are missing in the dataset.

\section{Methods}
Our main goal is to build a model that can recognize the grapheme components: \emph{roots}, \emph{vowels}, and \emph{consonants} from a given Bengali grapheme image with high accuracy. As already discussed in the \emph{Related work} section, the MLP does not provide better performance to recognize handwritten bengali alphabet so to validate it we use a simple MLP model without focusing much on its architecture. Later, we try to explore various deep learning techniques. We start with ResNet, after that we propose our own model to compete in the Kaggle competition. Finally, we try RPN to recognize a single component from a given image that contains \emph{grapheme root}, \emph{consonant diacritic} and \emph{vowel diacritic}.

 \subsection{Multi-Layer Perceptron (MLP)}
\begin{figure}
     \centering
     \includegraphics[width=0.5\textwidth]{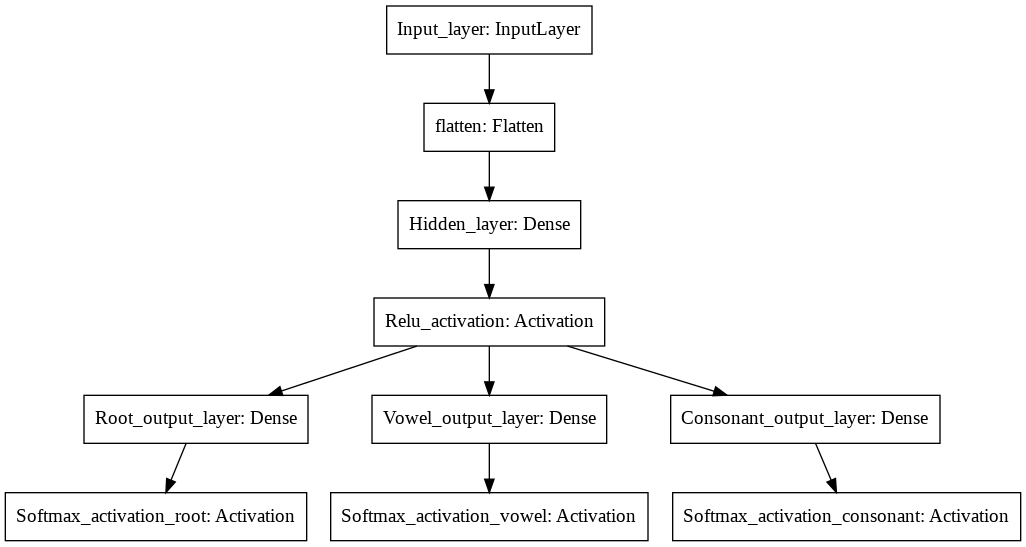}
     \caption{Multi-Layer Perceptron Model}
     \label{fig:mlp}
     \end{figure}

At first, we try to classify the components from a image using a simple Multi-Layer Perceptron (MLP) model. The proposed MLP model is shown in figure \ref{fig:mlp}. A single hidden layer with 5120 hidden nodes is used. We use \emph{relu} activation function in the hidden layer. In the output layer, \emph{softmax} function is considered as the activation function. We use \emph{adam} optimizer and \emph{categorical\_crossentropy} as loss function. The preprocessing of the data set is described in our \emph{proposed method} section. As MLP is a simple model, it does not perform well while classifying grapheme components (shown in result section). So, we start to explore various deep learning techniques.

\subsection{ResNet}
While experimenting with different models we explore ResNet50 to train. Other than a few changes in the hyper-parameters the model architecture is the same. At first we try to find the model performance on the 4 subsets of data separately (shown in figure 12). However, the training time is 4 hours which exceeds the time limit (2 hours) of the competition. Therefore, we move forward to design our own model that can achieve high accuracy while meeting the time limit constraint. To train the model we use \emph{adam} optimizer and \emph{categorical\_crossentropy}.

\begin{figure*}[ht]
\vskip 0.2in
\begin{center}
\centerline {
\begin{tabular}{ccc}
  \includegraphics[width=0.3\textwidth]{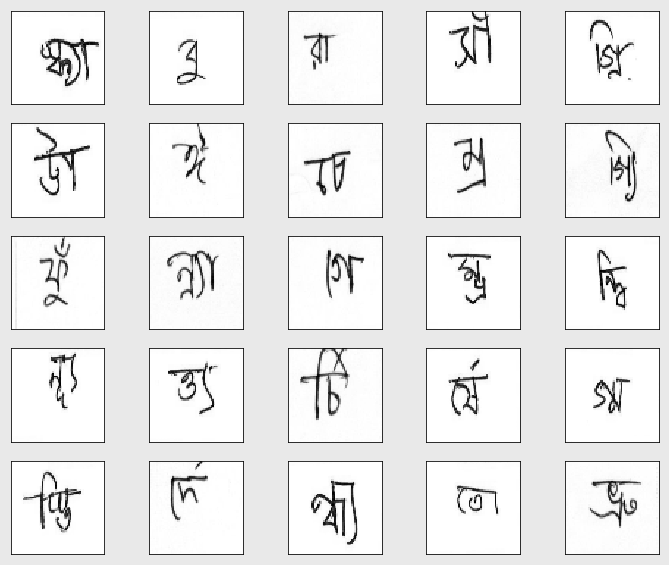} &   \includegraphics[width=0.3\textwidth]{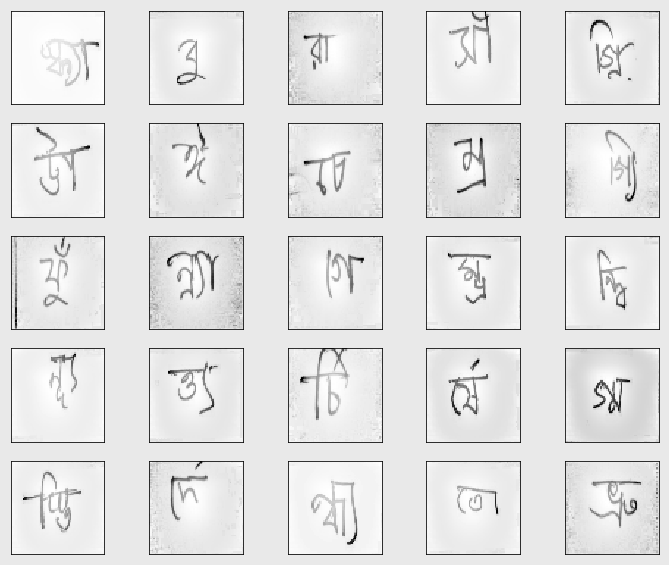} & \includegraphics[width=0.3\textwidth]{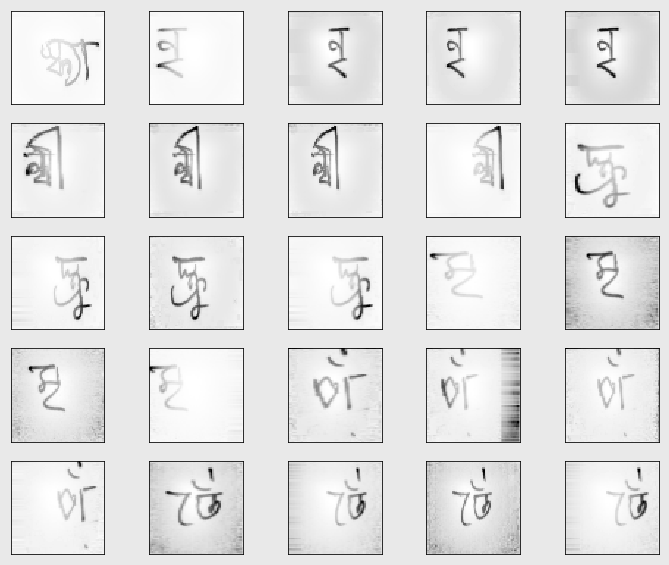}\\
(a) Original dataset & (b) Preprocessed dataset  & (c) Augmented dataset\\[6pt]
\end{tabular}
}

\caption{Overview of original dataset, preprocessed dataset, and augmented dataset}
\label{dataset-overview}
\end{center}
\vskip -0.2in
\end{figure*}

   \subsection{Proposed Method}
   We propose a convolution neural network (CNN) model to classify grapheme components: \emph{roots}, \emph{vowels}, and \emph{consonants} from a given Bengali grapheme image. The missing combinations of graphmeme components in the dataset, the high number of classes for each component, and the huge size of dataset i.e. $\approx4.83$ GB make this classification task very challenging. In addition to that, the model needs to finish it's execution within $2$ hours limit (Kaggle kernel requirement). Given those challenges, we try a set of different models from shallow to deep, from wider to deeper model and finally, we come up with a model that fits the dataset best and gives good accuracy when satisfying Kaggle notebook requirement as well.
   
   \subsubsection{Preprocessing}
   The dataset consists of $0-255$ value gray images that we convert to $0-1$ $float32$ image by dividing each pixel with $255.0$. We also resize the images to $64x64$ shape. Then we apply standardization on each image as $(x-mean)/std$ where $x$ is the input image, std is the standard deviation calculated from the whole dataset. Figure \ref{dataset-overview} $(b)$ shows a snippet of our preprocessed dataset. We also look into normalization; however, we confirm that standardization gives better performance over the normalization for this dataset.
   
   \subsubsection{Data Augmentation}
   We observe that there are different types of variances in the dataset for instance not all the images have regular shaped graphemes; some are slightly rotated; some have slightly different width or height scaling. Figure \ref{dataset-overview} $(a)$ shows some sense of our observed variances in the dataset. Therefore, we augment the dataset by applying Keras's $random \_rotation$, $random\_shift$,  $random \_shear$, and $random\_zoom$ function so that the dataset has a good distribution of these features.
   
   \subsubsection{Model}
\begin{figure*}[ht]
    \vskip 0.2in
    \begin{center}
    \centerline {
    \begin{tabular}{cc}
      \includegraphics[width=0.45\textwidth]{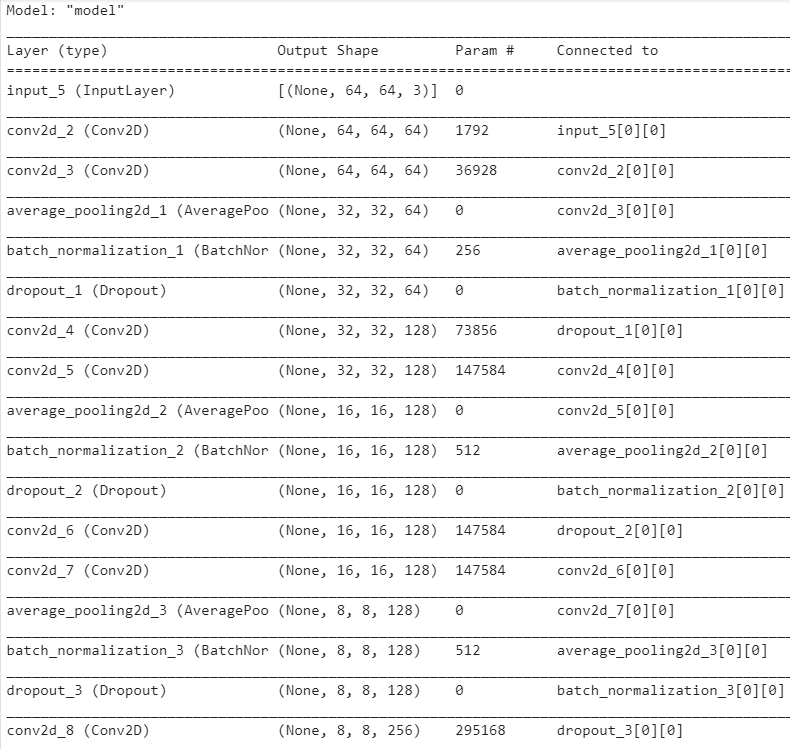}&
      \includegraphics[width=0.45\textwidth]{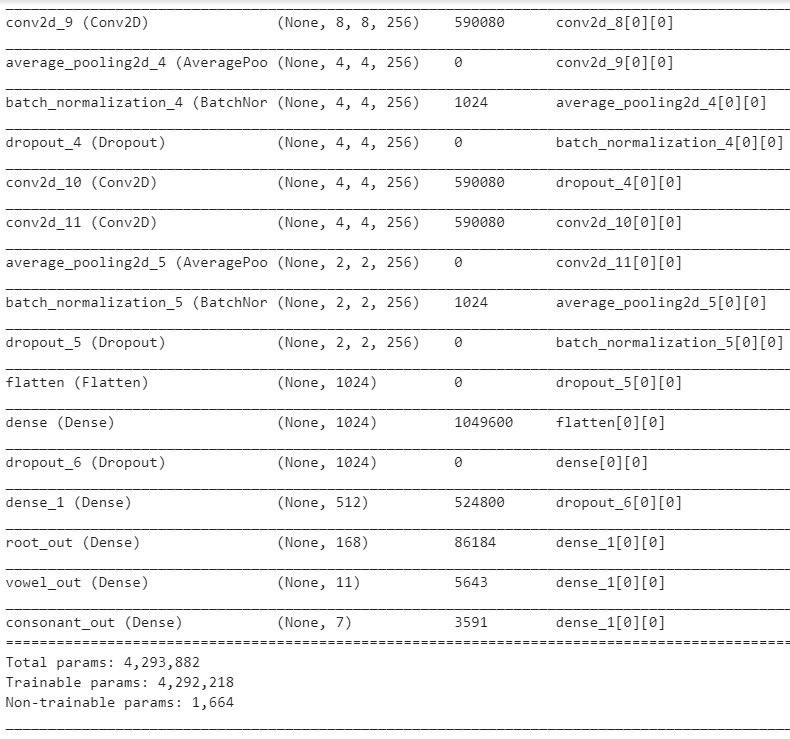}\\
    \end{tabular}
    }
    \caption{Summary of the proposed CNN model}
    \label{proposed-model-summary}
    \end{center}
    \vskip -0.2in
    \end{figure*}

\begin{figure*}[ht]
    \vskip 0.2in
    \begin{center}
    \centerline {
    \begin{tabular}{c}
      \includegraphics[width=0.7\textwidth]{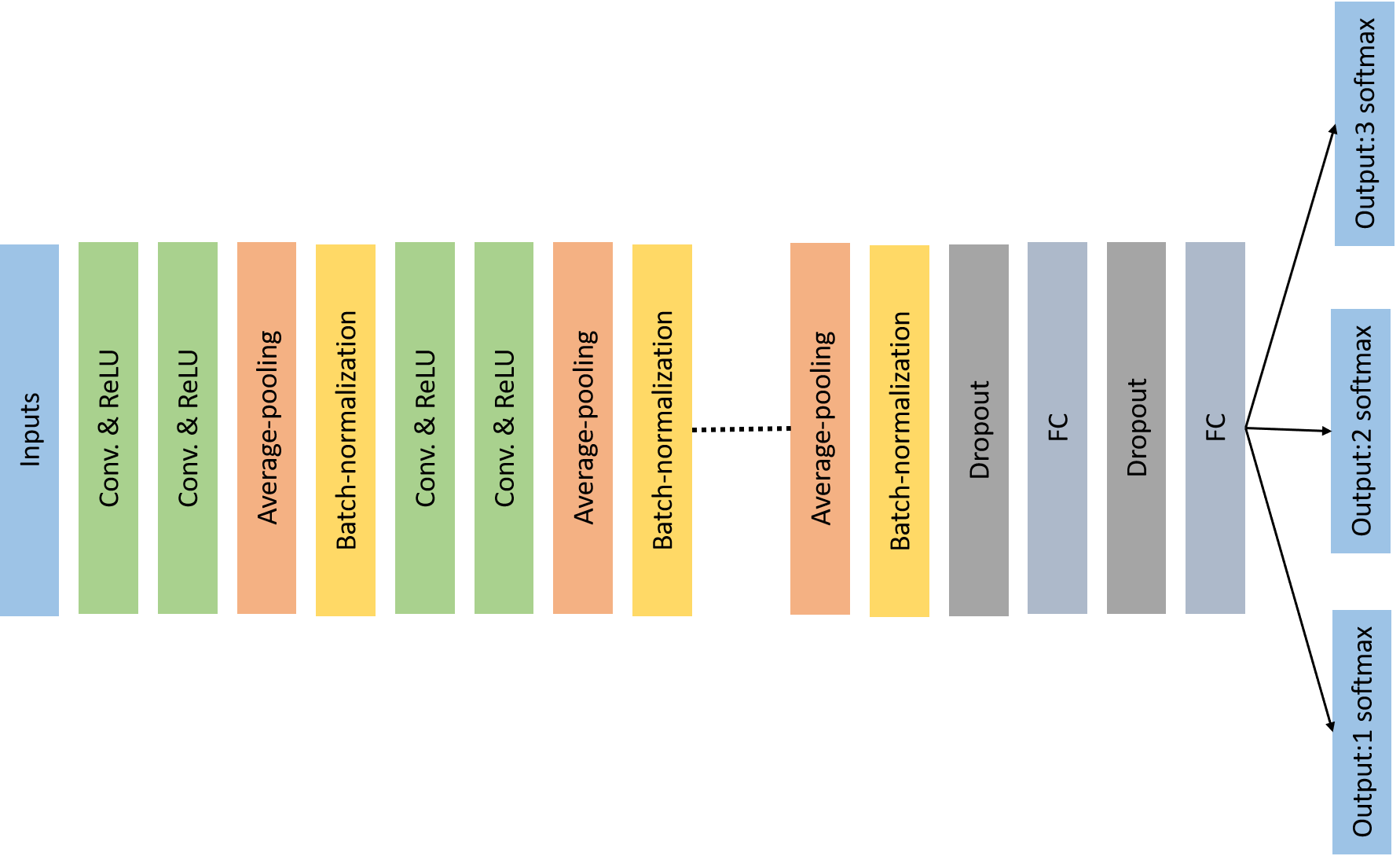}
    \end{tabular}
    }
    \caption{Architecture of the proposed CNN model}
    \label{proposed-model-architecture}
    \end{center}
    \vskip -0.2in
    \end{figure*}

   In our proposed model, we use total $11$ different convolution layers with each having $SAME$ padding and $relu$ activation function. Model inputs are $64x64$ images and outputs are defined as $[root,vowel,consonant]$ i.e. for a single input grapheme image, the model predicts $3$ outputs together for different components. In the output layer we use $softmax$ activation function. Throughout the model, We apply a bunch of $AveragePooling2D$ functions to reduce the input size to the next convolution layer. After each $AveragePooling2D$, we also apply $BatchNormalization$ with momentum $0.8$ to speed up the model training time as we have a $2hrs$ runtime restriction set by Kaggle. Furthermore, we increase the number of kernel filters as we go deeper so that the model can capture more variant features present in the input image. To avoid overfitting, we further use $Dropout$ throughout the model. We choose dropout value $0.4$ that gives us smooth learning without any overfitting. Our model uses $Adam$ optimizer and $sparse\_categorical\_crossentropy$ as a loss function. The total trainable parameters is $4,292,218$ and non-trainable parameters is $1,664$. Figure \ref{proposed-model-summary} shows the summary of the proposed model and Figure \ref{proposed-model-architecture} shows the overall architecture of our model.
   
\subsection{Region Proposal Network (RPN) using VGGNet}
\begin{figure}
    \vskip 0.2in
    \begin{center}
    \centerline {
    \begin{tabular}{cc}
     \includegraphics[width=0.25\textwidth]{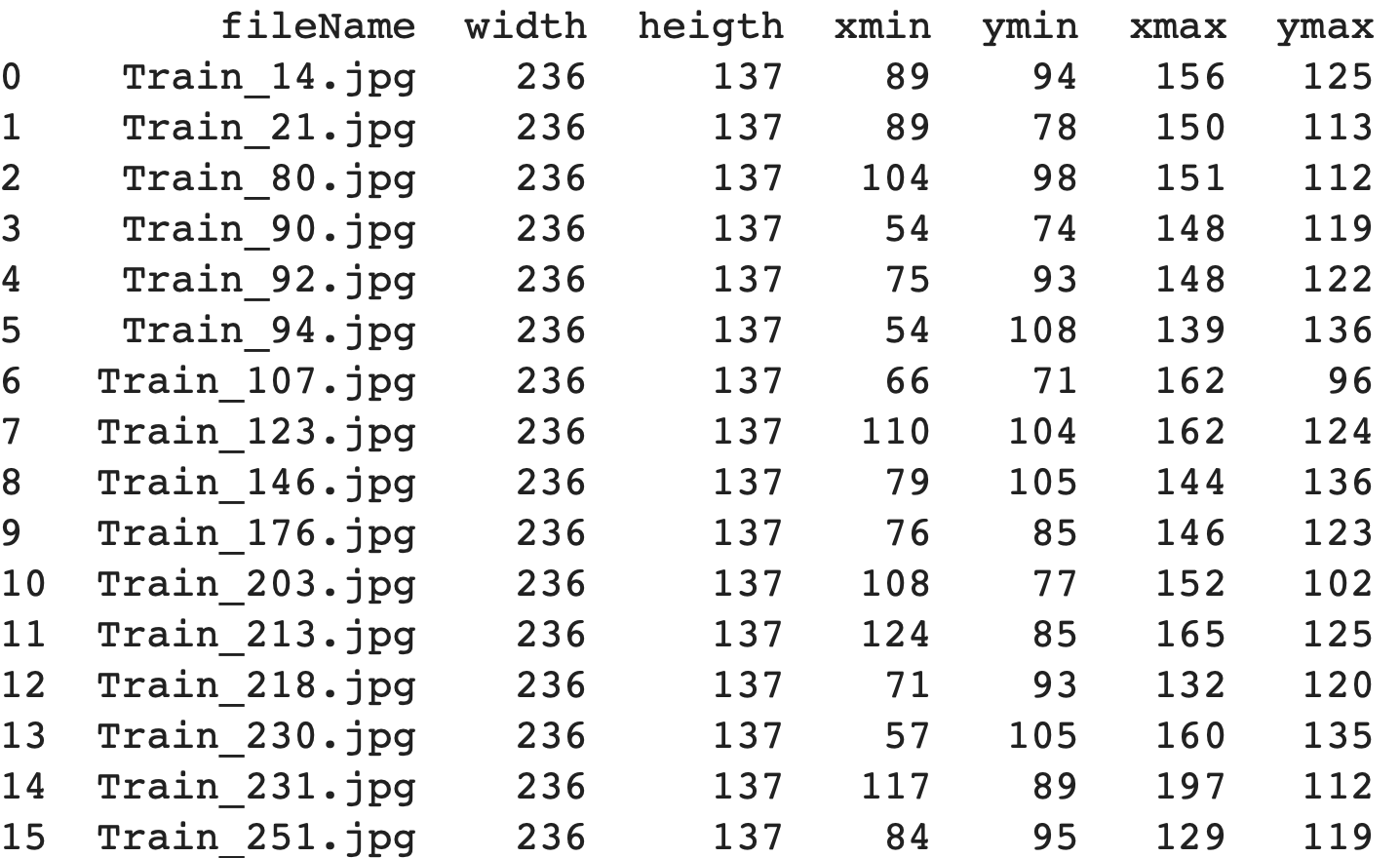}&
    
    \includegraphics[width=0.25\textwidth]{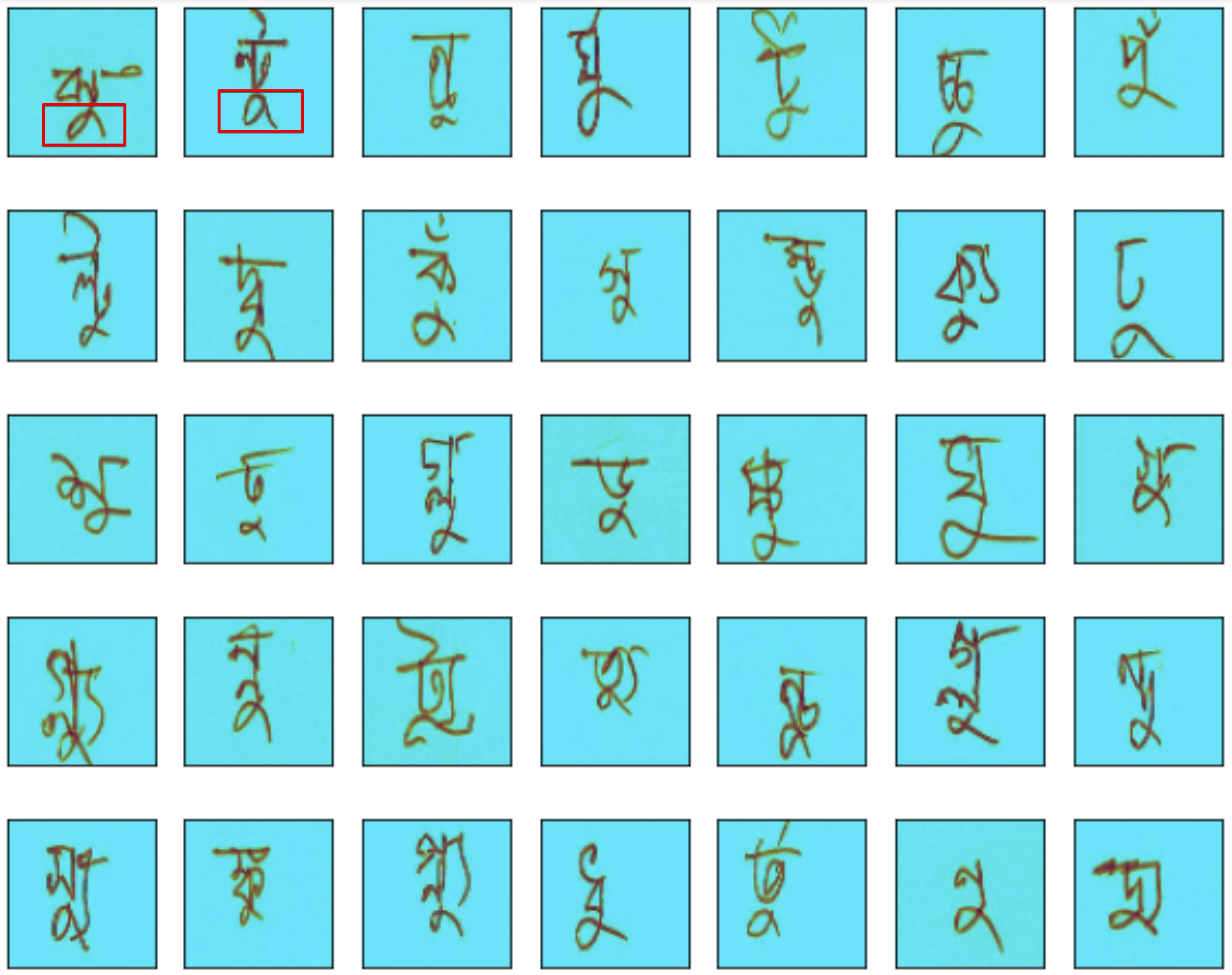}\\
      (a) Input file for RPN & (b) a portion of the images\\
    \end{tabular}
    }
    \caption{Images used in RPN}
    \label{rpn_image}
    \end{center}
    \vskip -0.2in
    \end{figure}

\begin{figure}
     \centering
     \includegraphics[width=0.4\textwidth]{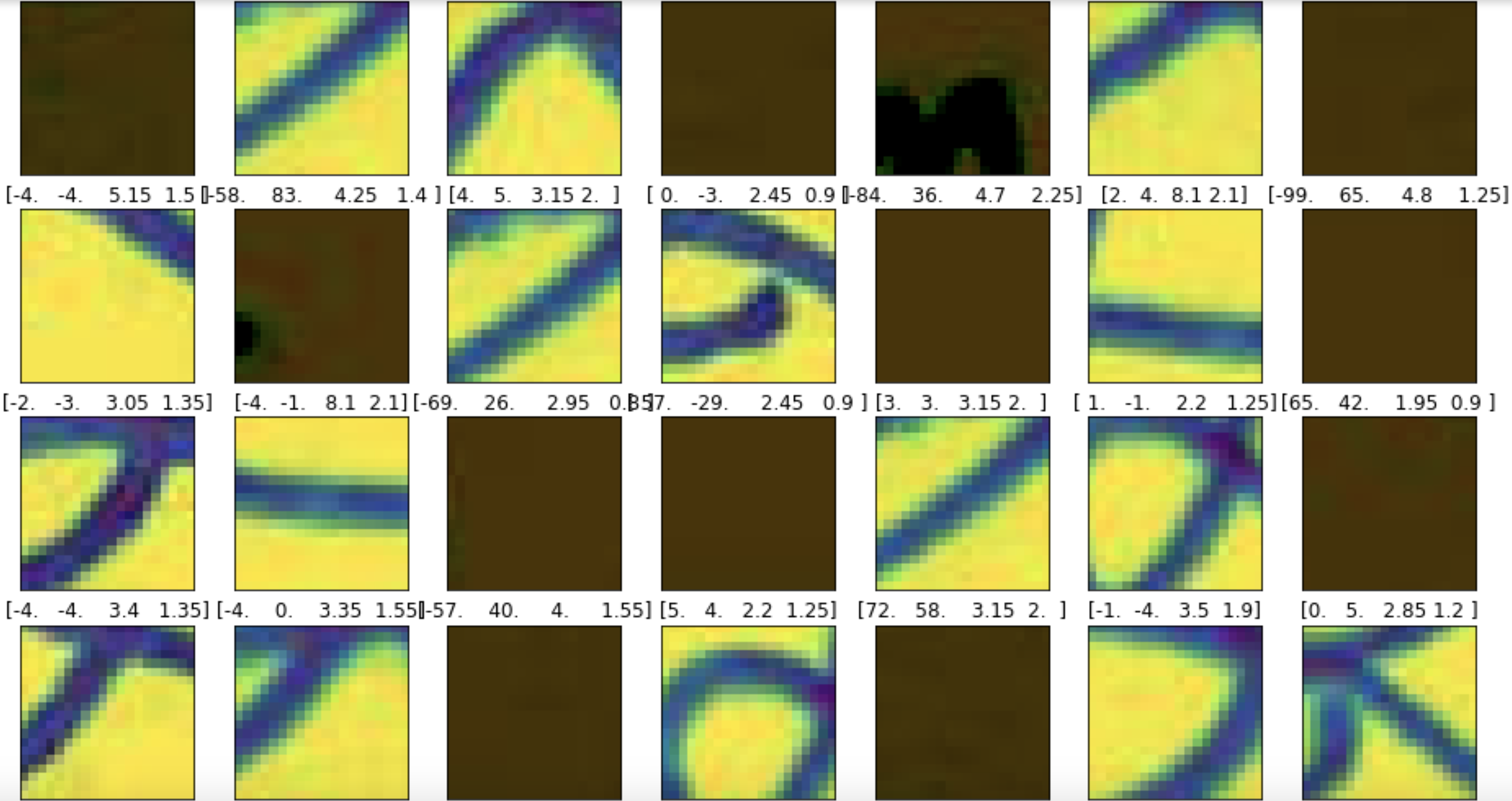}
     \caption{Images after sampling}
     \label{fig:samples_rpn}
     \end{figure}

To identify the \emph{root}, \emph{vowel diacritic} and \emph{consonant diacritic} from a given image, we also try to use Regional Proposal Network (RPN) \cite{rpnfaster}  method. In this project, we try to use RPN to identify a specific \emph{vowel diacritic} from a given image. So, we start with a very small number of images (42 images with a fixed \emph{vowel diacritic}) from the dataset. A portion of the images are shown in figure\ref{rpn_image}(b). The vowel diacritic is focused using red colored rectangle in the first two images. We also calculate the $x_{min}, y_{min}, x_{max}, y_{max}$ of the vowel diacritic and put those in the input file (figure \ref{rpn_image}(a)). Then we generate sample images using 20 $\times$ 20 anchor box. A portion of generated images are shown in figure \ref{fig:samples_rpn}. After that, VGG16 (up to \emph{block3\_conv3 layer}) \cite{simonyan2014deep} is used to predict the \emph{vowel diacritic}. \emph{adam} optimizer and a custom loss function (combination of \emph{class} loss and \emph{regression} loss) is considered. The model produces some false positives as well. To remove the false positives, we only keep those outputs whose bounding box area intersect with the vowel diacritic area (shown in figure \ref{fig:detection_rpn}). Recognizing all the three components from an image can be a potential future direction.
    
\section{Results and Discussion}

\subsection{MLP}
\begin{figure*}[ht]
\vskip 0.2in
\begin{center}
\centerline {
\begin{tabular}{ccc}
  \includegraphics[width=0.3\textwidth]{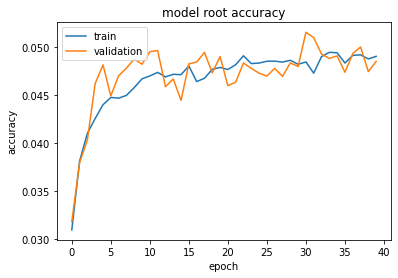} &   \includegraphics[width=0.3\textwidth]{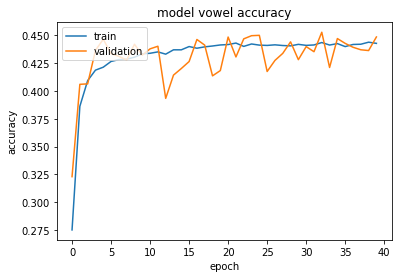} & \includegraphics[width=0.3\textwidth]{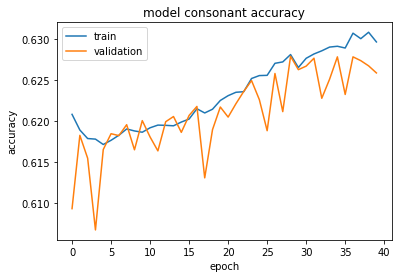}\\
(a) Grapheme Root Accuracy & (b) Vowel Diacritics Accuracy  & (c) Consonant Diacritics Accuracy\\[6pt]
\end{tabular}
}

\caption{Accuracy using Multi-Layer Perceptron }
\label{mlp_accuracy}
\end{center}
\vskip -0.2in
\end{figure*}

To train the model, the epoch, batch size and validation split are considered 40, 32, and 0.3 respectively. The model is trained on 100000 dataset which is half of the total dataset. The accuracy achieved by MLP is shown in figure \ref{mlp_accuracy}. Since the graphemes of Bengali language are complex in nature, the performance of the MLP is not so good. The grapheme root has more classes and is more complex than other two diacritics so identifying the grapheme root is more challenging than identifying the other two components. Figure \ref{mlp_accuracy} also validates it. 

\subsection{ResNet}
\begin{figure*}
\vskip 0.2in
\begin{center}
\centerline {
\begin{tabular}{ccc}
  \includegraphics[width=0.3\textwidth]{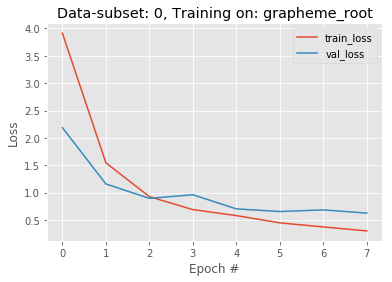} &   \includegraphics[width=0.3\textwidth]{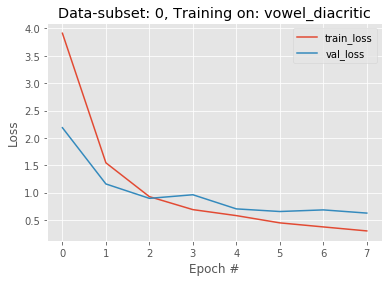} & \includegraphics[width=0.3\textwidth]{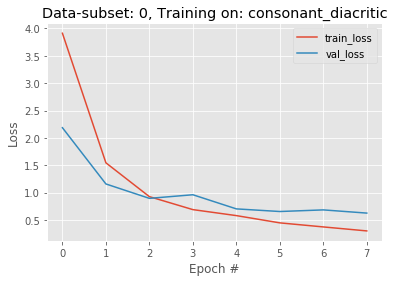}\\
(a) Graphem roots & (b) Vowel diacritics  & (c) Consonant diacritics\\[3pt]
  \includegraphics[width=0.3\textwidth]{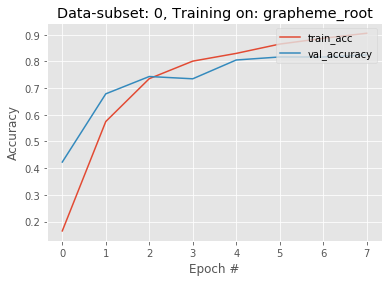} &   \includegraphics[width=0.3\textwidth]{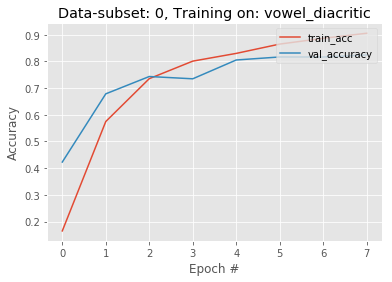} & \includegraphics[width=0.3\textwidth]{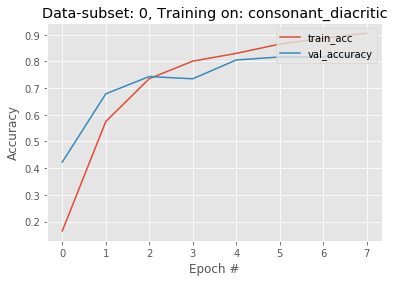}\\
(d) Graphem roots & (e) Vowel diacritics  & (f) Consonant diacritics\\[6pt]
\end{tabular}
}
\caption{Performance of the ResNet50 on one subset}
\label{fig:res50}
\end{center}
\vskip -0.2in
\end{figure*}

Performance of the ResNet50 on one of the four data subsets are shown in figure \ref{fig:res50}. Epoch and batch size are considered 20 and 128 respectively. The dataset split into 80\% and 20\% for training and validation. The accuracy of the train and validation should follow the same trend over time. 

   \subsection{Proposed Model}
\begin{figure*}[ht]
    \vskip 0.2in
    \begin{center}
    \centerline {
    \begin{tabular}{ccc}
      \includegraphics[width=0.3\textwidth]{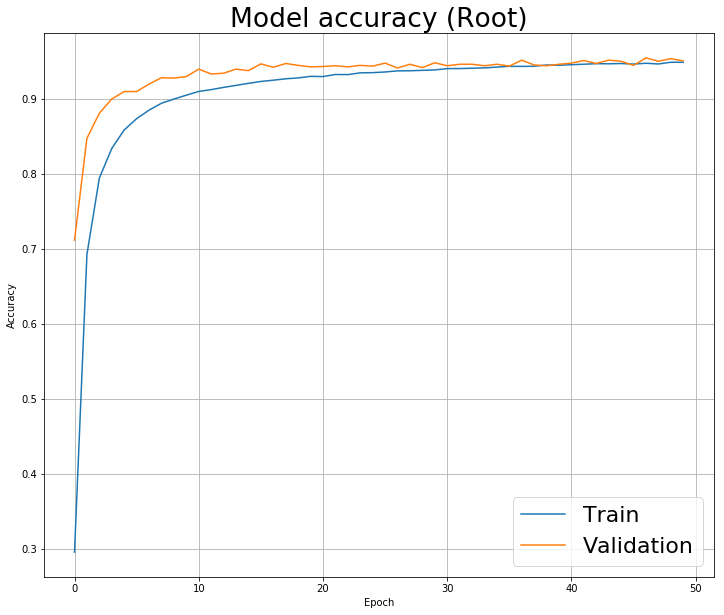}&
      \includegraphics[width=0.3\textwidth]{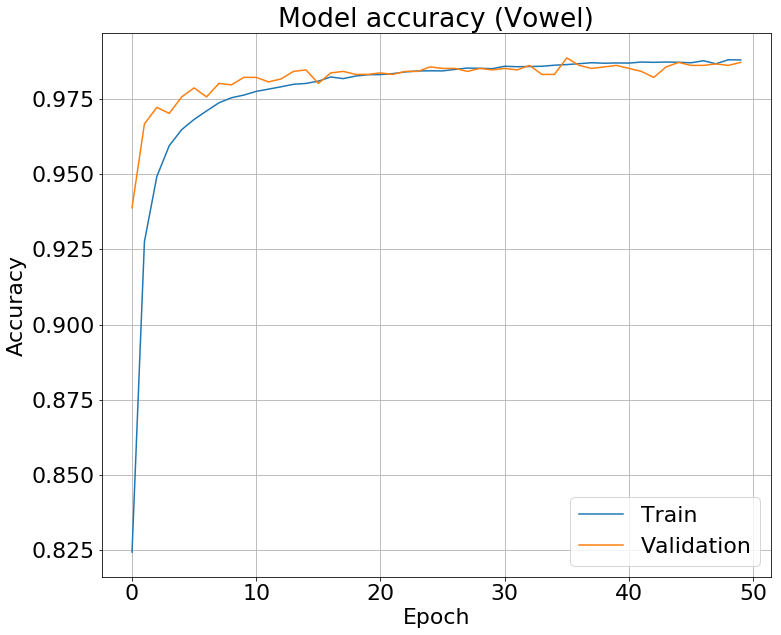} &
      \includegraphics[width=0.3\textwidth]{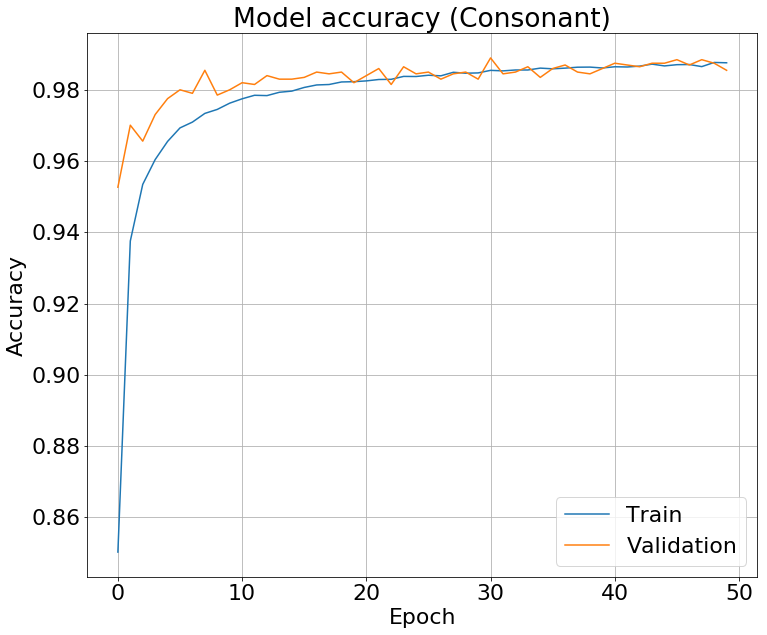}\\
      (a) Train vs validation accuracy (root) & 
      (b) Train vs validation accuracy (vowel) &
      (c) Train vs validation accuracy (consonant)
      \\ [6pt]
      \includegraphics[width=0.3\textwidth]{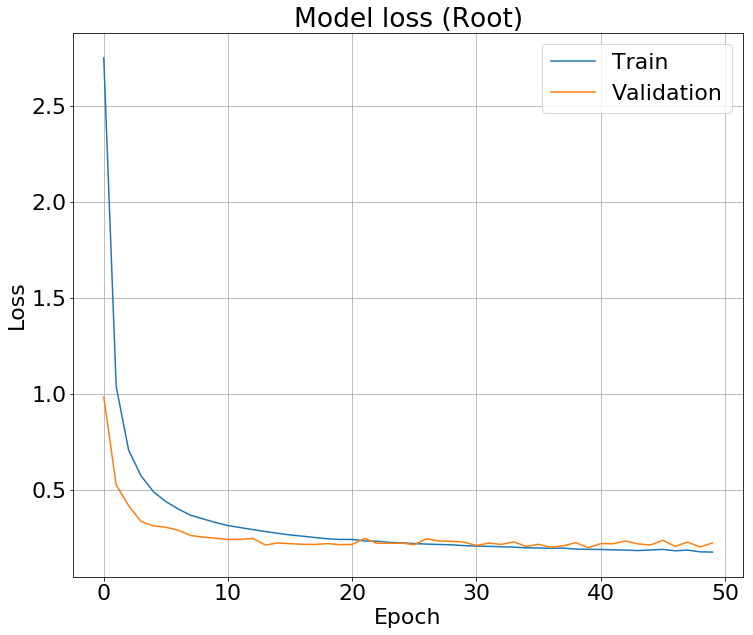}&
      \includegraphics[width=0.3\textwidth]{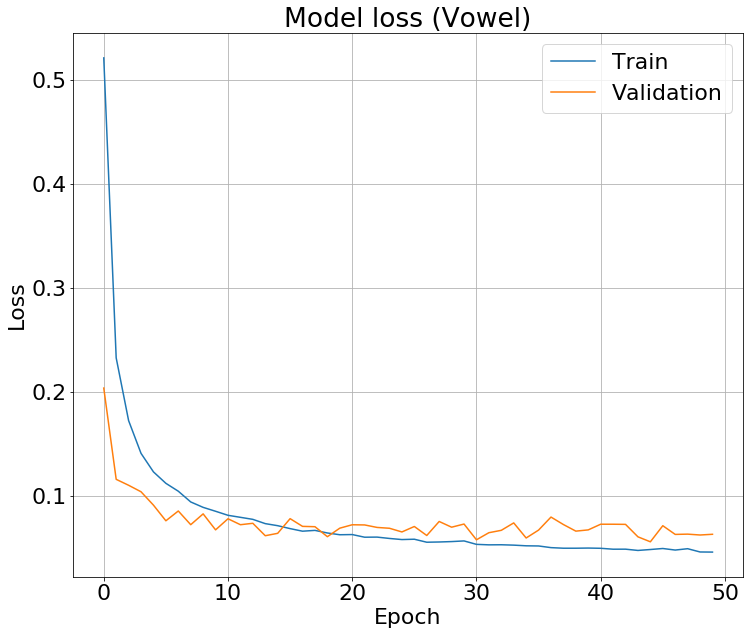} &
      \includegraphics[width=0.3\textwidth]{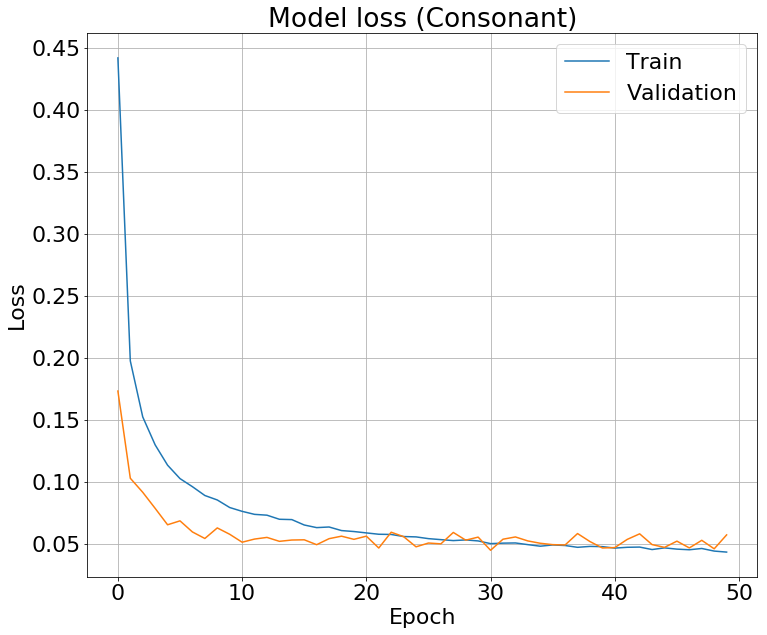}\\
      (d) Train vs validation accuracy (root) & 
      (e) Train vs validation accuracy (vowel) &
      (f) Train vs validation accuracy (consonant)
      \\
    \end{tabular}
    }
    \caption{Train vs Validation accuracy/loss of proposed CNN model}
    \label{proposed-model-training}
    \end{center}
    \vskip -0.2in
    \end{figure*}

    In methods, we described our best and most recent model. However, we additionally test different models and submitted to Kaggle. Table \ref{kaggle-results} shows the different submission results we achieve. In rest of the section, we will refer to only our most recent and best model. 
    We train our model for $50$ epoch using batch size $64$ and tuned dropout parameter as $0.4$ based on the over-fitting we observe during the training process. Figure \ref{proposed-model-training} shows how our model is fitted to training data and validation data while the training process is ongoing. It demonstrates that the proposed model learns quickly in the first few epoch because we use batch normalization to speed up the learning to satisfy Kaggle's runtime requirement. Then the training and validation loss get saturated gradually. Table \ref{table-accuracy-comparison} shows, how the proposed CNN model outperforms other models in terms of accuracy.
    
    \begin{table}[ht]
    \caption{Kaggle submission results; public score on 54\% data; private score is only disclosed after the competition ends}
    \label{kaggle-results}
    \vskip 0.15in
    \begin{center}
    \begin{small}
    \begin{tabular}{l c c c}
      Submission & Status & Private score & Public score \\
      \hline \
     1 & Succeeded & 0.8829 & 0.9451\\
     2 & Succeeded & 0.8795 & 0.9401\\
     3 & Succeeded & 0.8799 & 0.9388\\
     4 & Notebook Exceeded & X & X\\
     5 & Succeeded & 0.8812 & 0.9437\\
     6 & Succeeded & 0.8786 & 0.9416\\
     7 & Succeeded & 0.8834 & 0.9500\\
    \end{tabular}
    \end{small}
    \end{center}
    \vskip -0.1in
    \end{table}
    
    \begin{table*}[ht]
    \caption{Comparison of different models in terms of accuracy (train/validation)}
    \label{table-accuracy-comparison}
    \vskip 0.1in
    \begin{center}
    \begin{small}
    \begin{tabular}{lrrr}
      Model & Root accuracy &  Vowel accuracy & Consonant accuracy \\
      \hline
     MLP & $4.95\%/5.16\%$ & $44.37\%/45.26\%$ & $63.08\%/62.79\%$\\
     Resnet & $90.45 \%/91.50\%$ & $97.67\%/96.35\%$ & $98.23\%/96.81\%$\\
     Proposed CNN & $94.85\%/95.32\%$ & $98.79\%/98.61\%$ & $98.78\%/98.76\%$ \\
     RPN & NA & NA & NA\\
    \end{tabular}
    \end{small}
    \end{center}
    \vskip -0.1in
    \end{table*}

\subsection{RPN using VGGnet}
\begin{figure}
     \centering
     \includegraphics[width=0.4\textwidth]{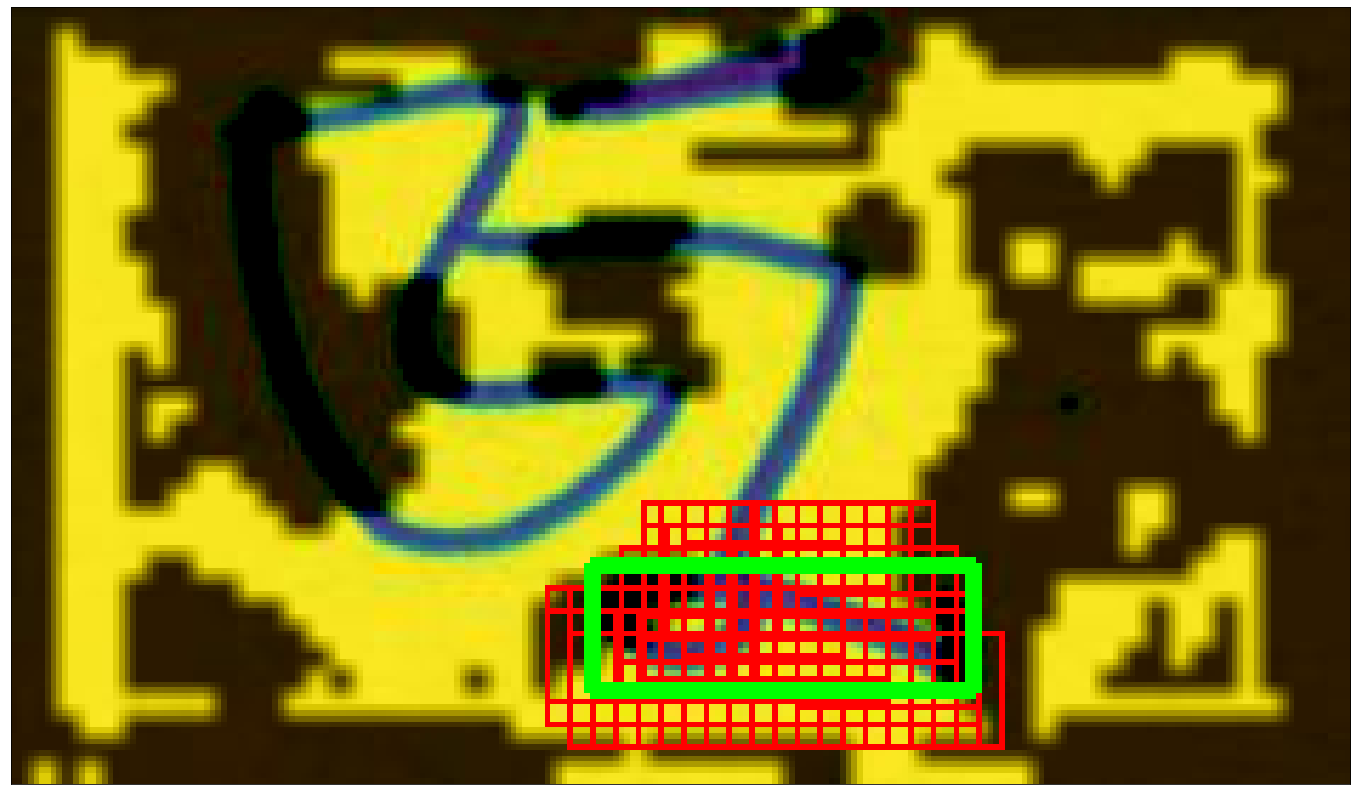}
     \caption{Object detection using RPN}
     \label{fig:detection_rpn}
     \end{figure}

    In RPN model, epoch and batch size are considered 500 and 128 respectively.
    Figure \ref{fig:detection_rpn} shows that RPN can detect a specific object (\emph{vowel diacritic} shown using red colored bounding boxes) among various other different objects (\emph{grapheme roots}, \emph{consonant diacritic}) in an image.  
    
\section{Conclusion} 
Though CNN represents a huge breakthrough in image classification challenges recently, it is not being explored widely for the Bengali grapheme classification. In this paper, we propose a new CNN model to classify the different components of Bengali grapheme - \emph{root}, \emph{vowel}, and \emph{consonant} with high accuracy. Our model provides validation root accuracy $95.32\%$, vowel accuracy $98.61\%$, and consonant accuracy $98.76\%$. We also explore the performance and limitation of existing models - Multi-Layer Perceptron (MLP) and ResNet50 and give a comparison with our model. We also study and explore Region Proposal Network (RPN) using VGGNet with a limited setting.   








\bibliographystyle{./IEEEtran}
\bibliography{./IEEEexample}

\vspace{12pt}

\end{document}